%% file: main.tex
\documentclass[11pt]{article}

\usepackage[final]{acl}

\usepackage{times}
\usepackage{latexsym}

\usepackage[T1]{fontenc}

\usepackage[utf8]{inputenc}

\usepackage{microtype}

\usepackage{inconsolata}

\usepackage{graphicx}
\include{lib_by_dipta}

\author{
    Shubhashis Roy Dipta\hspace{.1em}$^{\,1}$,\quad
    Khairul Mahbub\hspace{.1em}$^{\,2}$,\quad
    Nadia Najjar\hspace{.1em}$^{\,2}$ \vspace{2mm}\\
    $^1$\textbf{University of Maryland, Baltimore County} \\
    $^2$\textbf{University of North Carolina at Charlotte}\vspace{2mm}\\
    {\texttt{sroydip1@umbc.edu}\quad \texttt{kmahbub@charlotte.edu}}
}

\usepackage{xspace}
\newcommand{\ours}{\textsc{GanitLLM}\xspace} %
\newcommand{\ourdb}{\textsc{Ganit}\xspace}
\newcommand{\ourdev}{\textsc{Ganit-Dev}\xspace}
\newcommand{\ourtrain}{\textsc{Ganit-Train}\xspace}
\newcommand{\oursft}{\textsc{Ganit-SFT}\xspace}
\newcommand{\ourrlvr}{\textsc{Ganit-RLVR}\xspace}
\newcommand{\cgrpo}{\textsc{Curriculum-GRPO}\xspace}

\title{\ours: Difficulty-Aware Bengali Mathematical Reasoning through \cgrpo}

\begin{document}
\maketitle

\input{sections/00_abstract}
\input{sections/01_introduction}

\input{sections/90_related_works}

\input{sections/02_Method}

\input{sections/03_results}

\input{sections/appendix/grpo_vs_cgrpo}

\input{sections/appendix/qualitative_analysis}

\input{sections/91_conclusion}

\input{sections/99_limitations}

\section*{Acknowledgment}
Some experiments were conducted on the UMBC HPCF, supported by the National Science Foundation under Grant No. CNS-1920079. %
This material is also based on research that is in part supported by DARPA for the SciFy program under agreement number HR00112520301. The U.S. Government is authorized to reproduce and distribute reprints for Governmental purposes notwithstanding any copyright notation thereon. The views and conclusions contained herein are those of the authors and should not be interpreted as necessarily representing the official policies or endorsements, either express or implied, of DARPA or the U.S. Government.

\bibliography{normalized}

\clearpage
\appendix
\input{sections/appendix/100_apendix}

\end{document}

%% file: lib_by_dipta.tex
\usepackage[shortlabels,inline]{enumitem}
\usepackage{tikz,lipsum}
\usepackage[most]{tcolorbox}
\usepackage{ragged2e}
\usepackage[dvipsnames]{xcolor}
\usepackage{amsmath}
\usepackage{booktabs}
\usepackage{tabularray}
\usepackage{arydshln}
\usepackage{stmaryrd}
\usepackage{marvosym}
\usepackage{colortbl}
\usepackage{multicol}
\usepackage{multirow}
\usepackage{float}
\usepackage{makecell}
\usepackage{amssymb} %
\usepackage{algorithm} 
\usepackage{algpseudocode} 
\usepackage{algorithmicx} 
\usepackage{colortbl}
\usepackage{tablefootnote}

\usepackage{cleveref}
\crefname{table}{Table}{Tables}
\crefname{figure}{Fig.}{Figs.}
\crefname{algorithm}{Alg.}{}
\crefname{ALC@unique}{Line}{Lines}
\crefname{equation}{Eq.}{Eqs.}
\crefformat{section}{\S#2#1#3} 
\crefname{appendix}{App.}{Apps.}

\usepackage{soul}
\definecolor{tablegray}{RGB}{223, 242, 252}

\usepackage{todonotes}

\usepackage{minted}
\usepackage{fancyvrb}

\NewEnviron{prompt}[1][]{%
\begin{tcolorbox}[
    coltitle=white,
    colframe=black,
    colback=black!5!white,
    enhanced jigsaw,
    breakable,
    fontupper=\small,
    fontlower=\small,
    fonttitle=\small,
    title={#1}, %
]
\BODY
\end{tcolorbox}
}

%% file: sections/00_abstract.tex
\begin{abstract}
We present a Bengali mathematical reasoning model called \ours (named after the Bangla word for mathematics, \emph{Ganit}), together with a new difficulty-aware Bengali math corpus and a curriculum-based GRPO pipeline. Bengali is one of the world’s most widely spoken languages, yet existing LLMs either reason in English and then translate, or simply fail on multi-step Bengali math, in part because reinforcement learning recipes are tuned for high-resource languages and collapse under reward sparsity in low-resource settings. To address this, we construct \textbf{\ourdb}, a rigorously filtered and decontaminated Bengali math dataset with automatic difficulty tags derived from the pass@k of a strong evaluator model.
Building on this dataset, we propose \textbf{\cgrpo}, which combines multi-stage training (SFT + GRPO) with difficulty-aware sampling and verifiable rewards for format, numerical correctness, and Bengali reasoning. 
On Bn-MGSM and Bn-MSVAMP, \ours-4B improves over its Qwen3-4B base by \textbf{+8} and \textbf{+6} accuracy points, respectively, while increasing the percentage of Bengali reasoning tokens from \textbf{14\%} to over \textbf{88\%} and reducing average solution length from \textbf{943} to \textbf{193} words.
\footnote{\url{https://dipta007.github.io/GanitLLM/}}
\end{abstract}

%% file: sections/01_introduction.tex
\section{Introduction}
\input{tables/fig_intro}

Recent Large Language Models (LLMs) show strong reasoning performance across high-resource languages like English \cite{shi2022language}. In contrast, progress in low-resource languages remains limited \cite{lai2024mcot}. Bengali, the seventh most spoken language worldwide\footnote{\href{https://www.icls.edu/blog/most-spoken-languages-in-the-world}{\texttt{icls/most-spoken-languages-in-the-world}}}, clearly reflects this gap \cite{bhowmik2025evaluating}. Early efforts such as BanglaBERT \cite{bhattacharjee2021banglabert}, BanglaGPT \cite{salim2023banglagpt}, TituLLM \cite{nahin2025titullms}, and TigerLLM \cite{raihan2025tigerllm} tried to address this challenge. Yet progress in complex multi-step reasoning tasks in Bengali, particularly in mathematics, continues to lag due to the scarcity of high-quality Bengali mathematical reasoning datasets \cite{bhowmik2025evaluating}. This underscores the need for research on Bengali mathematical reasoning to advance low-resource LLMs and expand their applications for Bengali speakers.

While previous works have shown promising results in Bengali mathematics \citep{lai2024mcot}, they have only evaluated the final accuracy. 
\citet{shi2022language} found that intermediate reasoning in English leads to better reasoning performance (left in \cref{fig:intro}), but in our case, we want a model that not only answers correctly but also reasons in Bengali (right in \cref{fig:intro}) for end-user interpretability and understanding. Most mathematical LLM users, e.g., students, seek not only the answer but also the step-by-step reasoning to understand the solving process.
Additionally, we identified that traditional RL training recipes, e.g., GRPO even with a high-quality dataset fails to solve this problem due to the high dominance of high-resource languages in pre-training. \citet{wu2025confucius3} have shown that only GRPO training is enough to improve the mathematics capability in Chinese but \citet{chowdhery_palm_2022} revealed that the number of Bengali tokens in pre-training is $\sim$15 times fewer than Chinese (0.026\% vs 0.4\%), which makes Bengali a far rarer and harder to improve than Chinese.

We define the cold-start problem in GRPO training as the scenario in which the policy model—due to its limited capability in the target low-resource language—fails to produce any correct solutions within a rollout group. This results in zero rewards across all samples and, consequently, zero gradients. Such cases lead to highly inefficient training (see \cref{sec:impact_cgrpo}), posing a critical challenge in low-resource settings.
In this work, we introduce \textbf{(i) a difficulty-aware, rigorously-filtered and -processed high-quality Bengali math dataset, \ourdb}, and \textbf{(ii) a novel training recipe, \cgrpo to tackle the cold-start problem in low-resource languages}.
To the best of our knowledge, we are the first to develop a Bengali Math LLM that performs reasoning truly in Bengali, rather than translating \citep{shu_transcending_2024} or reasoning in English \citep{lai2024mcot}. To enable this, we construct a difficulty-tagged Bengali math dataset by combining and curating several existing high-quality Bengali math datasets and adopting the pass@k as the proxy difficulty score (\cref{sec:ganit}).
We categorize problems into Easy, Medium, Hard and Olympiad levels.
Next, we fine-tune the base-instruct model on our CoT-SFT variant to teach the model to reason in Bengali rather than English.
Finally, we apply Group Relative Policy Optimization (GRPO) \citep{shao2024deepseekmath} with three different reward scores (\cref{sec:ganitllm}): (i) Format, (ii) Accuracy, (iii) Bengali Reasoning. We have also modified the dataset sampling during training to tackle the cold-start problem in Bengali reasoning. We call this whole reinforcement learning procedure \cgrpo (\cref{sec:curriculum_grpo}). 

To summarize, our contributions are as follows:
\begin{enumerate}[leftmargin=\parindent, itemsep=0.1em, topsep=0.4em]
    \item We introduce a difficulty-tagged, rigorously-filtered Bengali Math dataset with verifiable answers.
    
    \item We introduce a novel GRPO training recipe, \textbf{\cgrpo}, which can effectively tackle the cold-start problem of low-resource languages during group-based rewarding.
    
    \item To the best of our knowledge, we release the \textbf{first Bengali Mathematical Reasoning model}. \ours outperforms models that are twice its size and achieves performance comparable to models four times larger, all while reasoning natively in Bengali rather than relying on English and using 79.5\% fewer tokens.
    \vspace{-2mm}
\end{enumerate}

\input{tables/all_dataset_stat}

%% file: tables/fig_intro.tex
\begin{figure}
    \centering
    \includegraphics[width=1.0\linewidth]{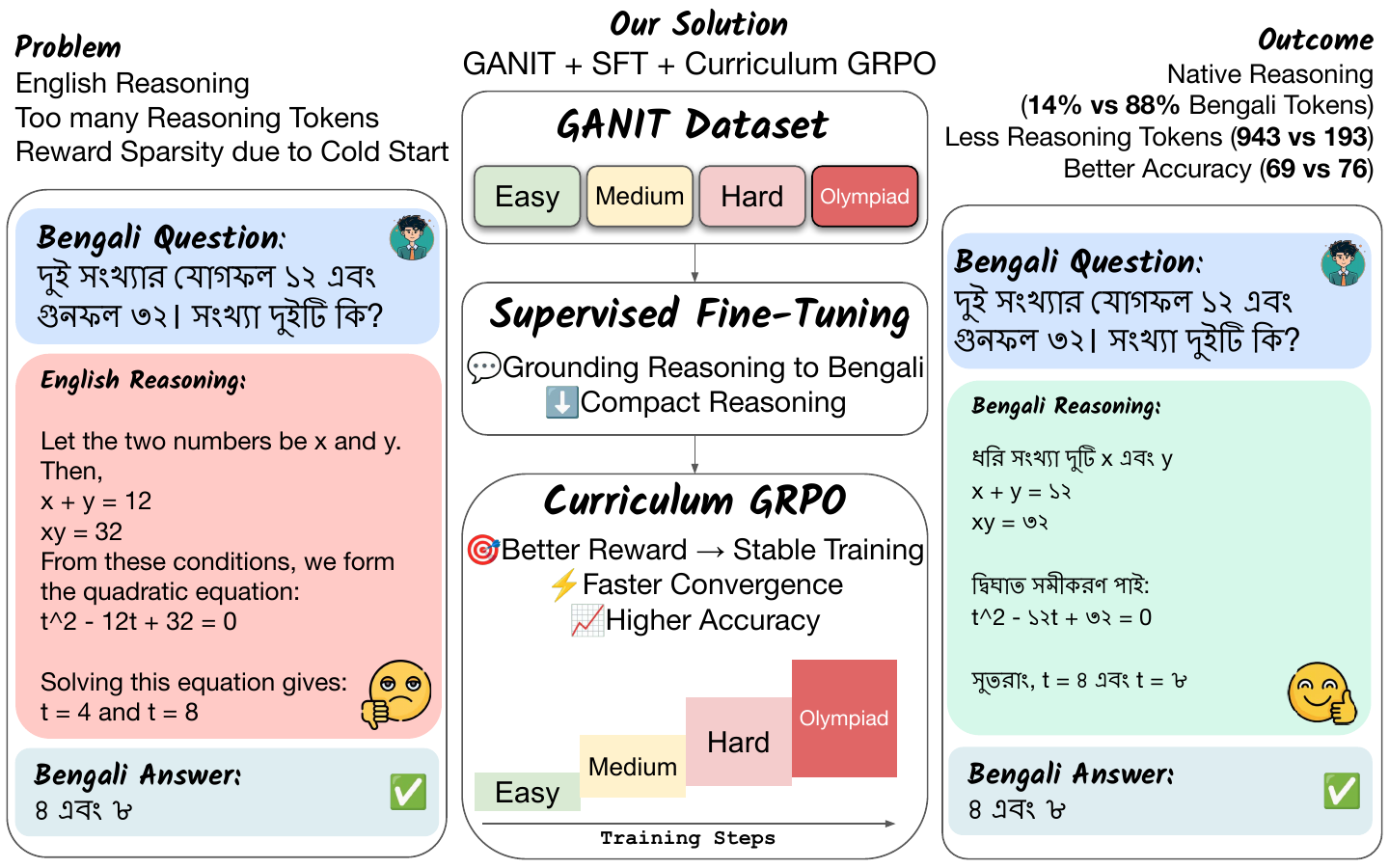}
    \caption{Overview of our approach for a Bengali mathematical reasoning model. (\textbf{Left}) Current models reason in English even for Bengali questions, resulting in reduced interpretability for native speakers. (\textbf{Center}) Our solution combines the \ourdb dataset with SFT to ground reasoning in Bengali, followed by \cgrpo for efficient RL training. (\textbf{Right}) Our approach achieves native Bengali reasoning (88\% Bengali vs. 14\%), reduces reasoning tokens by 79\%, and improves accuracy from 69 to 76.}
    \label{fig:intro}
    \vspace{-5mm}
\end{figure}

%% file: tables/all_dataset_stat.tex
\begin{table*}[!t]
    \centering
    \resizebox{\textwidth}{!}{
    \begin{tabular}{@{}lccccr@{}}
        \toprule
        \textbf{Dataset} &
        \textbf{Problem} &
        \textbf{Solution} &
        \textbf{\makecell{Source \\ (if translated)}} &
        \textbf{Size} &
        \textbf{\makecell[r]{Human \\ Evaluation}} \\
        \midrule
        mCoT-MATH-bn \cite{lai2024mcot} & GT & GT & mCoT-MATH \cite{lai2024mcot} & 580k & 100\% \\
        NuminaMath-CoT-bn \cite{numinamath_cot_bn_2024} & LT & LT & NuminaMath-CoT \cite{numina_math_datasets} & 859k & 97\% \\
        s1k-Bangla \cite{s1k_32_bangla_2024} & LT & LT & s1K-1.1 \cite{muennighoff2025s1} & 1k & 96\% \\
        DL Sprint 3.0 \cite{dlsprint3_olympiad_2024} & HA & HA & -- & 200 & 96\% \\
        SOMADHAN \cite{paul2025leveraging} & HT & HT & GSM8K \cite{cobbe2021training} & 8.7k & 96\% \\
        \midrule
        Shomikoron \cite{aurpa2024shomikoron} & HA & HA & -- & 3.4k & 90\% \\
        Bangla-Math \cite{kawchar85_bangla_math} & HA & LG & -- & 1.5k & 88\% \\
        PatiGonit \cite{era2024empowering} & HT & HT & MAWPS \cite{koncel-kedziorski_mawps_2016} & 10k & 85\% \\
        BMWP \cite{mondal2025bmwp} & HA & HA & -- & 8.6k & 77\% \\
        \bottomrule
    \end{tabular}
    }
    \scriptsize{HT = Human Translated, LT = LLM Translated, GT = Google Translated, HA = Human Annotated, and LG = LLM Generated}
    \vspace{-2.5mm}
    \caption{Overview of the quality and statistics of open-source datasets based on manual evaluation. \textbf{We used only the datasets with human evaluation greater than 95\% (top 5 rows).} Human Evaluation reports the percentage of samples with both correct problem statements and valid solutions out of the total sampled dataset.}

    \label{tab:datasets_q_a_status}
    \vspace{-3mm}
\end{table*}

%% file: sections/90_related_works.tex
\section{Related Works}
Large language models (LLMs) have recently made significant progress in complex reasoning tasks \citep{du2025ulorl,abdin2024phi} like mathematics \citep{wu2025confucius3, zhang2025lessons, nazi2026dag}, coding \citep{ceka2025does,halim2025study,el2025competitive} and commonsense  \citep{gawin2025navigating,wang2025deciphering, krause2024data} reasoning. The OpenAI o1 model \cite{openai2025o3o4mini} achieves state-of-the-art reasoning on complex, multi-step tasks. However, OpenAI has never open-sourced their training recipe. DeepSeek-R1 \cite{guo2025deepseek} marks a major advance in reasoning through novel reinforcement learning–based training methods. These high-performance LLMs excel in mathematical reasoning for high-resource languages, but their limited interpretability in the target language highlights the need for multilingual research.

\subsection{Multilingual Math Reasoning}
There has been growing exploration of multilingual contexts.
\citet{lai2024mcot} introduced mCoT, a 7B model with multilingual Chain-of-Thought tuning, achieving consistent reasoning across eleven languages on proprietary LLMs. Similarly, the MathCritique pipeline on ChatGLM3-32B enhances mathematical problem-solving while preserving language ability, outperforming larger LLMs \citep{xu2024chatglm}. Confucius3-Math \cite{wu2025confucius3}, a 14B open-source model for Chinese K-12 math, uses reinforcement learning with targeted entropy regularization to deliver state-of-the-art reasoning. \citet{muennighoff2025s1} developed s1-32B with budget forcing, showing that simple test-time scaling yields up to {27\%} gains over o1-preview on competition math. MindMerger \cite{huang2024mindmerger} integrates LLMs with multilingual models to augment cross-lingual reasoning, improving accuracy on MGSM \cite{shi2022language} by up to 8\% in low-resource settings without parameter updates. MathOctopus \citep{chen2023breaking}, trained on MGSM8K-Instruct with rejection sampling and parallel corpora, outperforms open-source models and ChatGPT in few-shot multilingual math reasoning. While studies have explored multilingual settings or targeted specific languages (e.g. Chinese), low-resource languages such as Bengali remain underexplored. Moreover, most of the multilingual reasoning methods only target the final accuracy, while the reasoning tokens are still in English.

\subsection{Bengali Math Reasoning}
Bengali LLMs reasoning research is still in a rudimentary stage. \citet{nahin2025titullms} presented TituLLMs, the first large pretrained Bengali LLMs and five benchmarking datasets, underscoring the challenges of language adaptation. TigerLLM \cite{raihan2025tigerllm} surpasses both open-source and proprietary models on standard benchmarks, setting a new baseline for Bengali reasoning. Research on Bengali mathematical reasoning has primarily concentrated on the development of high-quality datasets.
 \citet{aurpa2024shomikoron} and \citet{era2024empowering} introduced Shomikoron and PatiGonit, showing transformer models’ effectiveness on Bengali math problems.
\citet{mondal2025bmwp} contributed a large Bengali Math Word Problem (BMWP) dataset and showed strong operation prediction with deep neural networks. SOMADHAN \cite{paul2025leveraging}, a manually created step-by-step reasoning dataset, illustrates that Chain-of-Thought prompting improves proprietary LLMs on multi-step tasks. BEnQA \cite{shafayat-etal-2024-benqa}, a bilingual K-12 math dataset, indicates LLMs lag in Bengali math reasoning but improve when augmented with English translation prompts. Community translations of popular datasets like Numinamath \cite{numina_math_datasets} on Hugging Face and Kaggle broaden access, but ensuring cross-source alignment and quality remains challenging \citep{numinamath_cot_bn_2024,dlsprint3_olympiad_2024,s1k_32_bangla_2024}.

%% file: sections/02_Method.tex
\section{Creating \ourdb} \label{sec:ganit}
\input{tables/fig_data_processing}
We construct \textsc{Ganit}, a rigorously-processed, difficulty-aware Bengali math dataset comprising both \ourtrain and \ourdev sets. \ourtrain consists of two distinct splits: CoT-SFT variant and RLVR\footnote{Reinforcement Learning with Verifiable Rewards}, designed specifically for instruction tuning and reinforcement learning pipelines, respectively. 
Motivated by the limitations of existing Bengali evaluation benchmarks (details in \cref{sec:limit_mgsm_msvamp}), we additionally develop a hold-out, difficulty-aware dev set, \ourdev, to assess the capabilities of \ours. 
The entire dataset creation pipeline is illustrated in \cref{fig:data_preparation}.

\subsection{Data Collection}
Prior research has shown that LLMs can achieve superior performance when trained on high-quality and diverse data, even when the overall data volume is limited \citep{muennighoff2025s1, raihan2024mojobench}. However, obtaining such high-quality datasets for low-resource languages remains challenging, particularly in specialized domains such as mathematics \citep{chen2023breaking}. To address this gap, we processed a large Bengali mathematical dataset ($\sim$1.5M) by collecting publicly available Bengali math datasets spanning human-authored, human-translated, LLM-translated, and Google-translated sources. The datasets span mathematical skills from basic arithmetic \cite{mondal2025bmwp} to advanced competition-level problems \cite{dlsprint3_olympiad_2024}.
Additionally, they incorporate high-quality samples from research repositories \cite{paul2025leveraging} and community-contributed resources \cite{numinamath_cot_bn_2024}. This comprehensive coverage enables us to get a seed dataset with good coverage across different genres of mathematics.

\subsection{Data Filtering}
We applied a rigorous data filtering pipeline to the collected datasets listed in \cref{tab:datasets_q_a_status}, ensuring that the resulting data is high-quality, well-formatted, deduplicated, and decontaminated.

\paragraph{Quality Screening:} The problems and solutions in the datasets can be categorized as follows: (i) human-annotated, (ii) human-translated, (iii) LLM-generated, (iv) LLM-translated, and (v) Google-translated. Two evaluators manually evaluated randomly sampled subsets from all datasets (100 from each) to ensure rigorous quality screening. Manual evaluation results are reported in \cref{tab:datasets_q_a_status}.
The background and expertise of the evaluators are summarized in \cref{app:evaluator}.

As expected, the manual evaluation shows that human-annotated, human-translated, and LLM-translated datasets exhibit higher quality than LLM-generated synthetic datasets. After quality screening, only low-error datasets were retained (accuracy > 95\%), reducing the total size from $\sim$1.5M to $\sim$1.1M instances.

\paragraph{Rule-based Filtering:}
We applied rule-based filtering to the selected datasets to ensure consistency and verifiability. Specifically, (i) only solutions with numerical values were retained to allow for verifiable rewards, (ii) only problems containing at least 99\% Bengali characters were included, and (iii) multiple-choice questions were excluded.

\paragraph{Deduplication:}
We employed a two-stage deduplication process: (i) fuzzy string–based matching using normalized Levenshtein distance to detect near duplicates (3-gram with 70\% threshold), and (ii) MinHash-based similarity detection (200 hash size with 50\% threshold).

\paragraph{Decontamination:}
To prevent data leakage from evaluation benchmarks, we applied MinHash-based decontamination against MGSM \citep{shi2022language} and MSVAMP \citep{chen2023breaking}. Training instances with similarity above 50\% were removed. This ensured the training data was decontaminated, allowing reliable evaluation.

\subsection{Difficulty-aware \ourdb} \label{sec:diff_tagging}
Inspired by the pass@k \citep{chen2021evaluating} metric, we have used a similar strategy to estimate the difficulty level of each problem. First, we identify a strong general LLM based on its performance on Bn-MGSM \citep{shi2022language} and Bn-MSVAMP \citep{chen2023breaking} test sets. We choose these as they are the standard test sets for multilingual math evaluation. To that purpose, we have evaluated 8 open-source models ranging from 8B to 72B and identified that the \texttt{Qwen3-32B} performs the best (details in \cref{app:eval_mgsm_msvamp}). 
Next, we use \texttt{Qwen3-32B} to generate 32 independent solutions for each problem with a temperature of 0.7 to balance between diversity and correctness. A problem was retained only if the model successfully solved it at least once, thereby filtering out (i) potentially mislabeled noisy data, and (ii) instances likely unsolvable by smaller models during GRPO's group rollout. Finally, we uniformly categorize problems into four difficulty buckets based on the number of successful generations: \textbf{Olympiad (1--8), Hard (9--16), Medium (17--24), and Easy (25--32)}, enabling granular control over complexity levels.

\subsection{\ourdev}
\paragraph{Motivation:} \label{sec:limit_mgsm_msvamp}
To assess the difficulty of the standard MGSM and MSVAMP datasets, we applied the same difficulty tagging pipeline to their Bengali counterparts, Bn-MGSM and Bn-MSVAMP. The fine-grained statistics are presented in \cref{tab:eval_data_stat}. As the results indicate, the standard evaluation datasets in Bengali are relatively easy for current LLMs to solve. To construct a more robust difficulty-aware evaluation, we sampled 30 problems from each fine-grained difficulty bucket (1--32), resulting in a total of 960 examples ($30 \times 32$).

\begin{table}[!t]
    \centering
    \resizebox{\columnwidth}{!}{
    \begin{tabular}{@{}lccc@{}}
       \toprule
       \textbf{Difficulty}  & \textbf{MGSM} & \textbf{MSVAMP} & \textbf{\ourdev} \\
       \midrule
       Easy  & 77.50 & 86.00 & 28.74 \\
       Medium  & 16.40 & 8.40 & 26.03 \\
       Hard  & 3.60 & 3.20 & 24.31 \\
       Olympiad  & 2.50 & 2.40 & 21.26 \\
       \bottomrule
    \end{tabular}
    }
    \caption{Difficulty statistics (in \% of total data) of the Bn-MGSM \citep{shi2022language}, Bn-MSVAMP \citep{chen2023breaking} and \ourdev datasets.}
    \label{tab:eval_data_stat}
\end{table}

\paragraph{LLM-based Filtering:}
Furthermore, to ensure the quality of the dev set, we applied an LLM-based filtering procedure using three proprietary models: \textbf{\texttt{GPT-5-mini}}, \textbf{\texttt{Gemini-2.5-Flash}}, and \textbf{\texttt{Grok-4-Fast}}. Each model was prompted to solve each problem independently three times. Following the majority voting strategy, we select an answer if it appears in at least two out of three generations. Then we mark the answer as correct or wrong for each of the models. Finally, we retain only those problems that were correctly solved by all three models. This dual-stage filtering process ensures that the final set contains only high-quality, validated problems, minimizing the risk of noisy data. Notably, even after both filtering stages, the distribution of examples across difficulty buckets remained relatively balanced, as shown in \cref{tab:eval_data_stat}.

\subsection{\ourtrain}
In our training split, we specifically constructed two distinct datasets: (i) for instruction tuning (problem, CoT, solution), and (ii) for reinforcement learning (problem, verifiable answer). Since the instruction tuning data is primarily used to teach the LLM to reason in Bengali rather than to optimize for correctness, we hypothesize that SFT is less sensitive to data imbalance. Based on this assumption, we constructed a difficulty-balanced dataset for the RL split and utilized the remaining (imbalanced) portion for instruction tuning.

To achieve fine-grained difficulty balancing, we moved beyond the standard four coarse buckets and instead considered the exact number of correct generations (ranging from 1 to 32). We then randomly sampled an equal number of instances for each count, ensuring uniform representation. This was particularly important for the RL split, which is more susceptible to overfitting.
Full statistics of \ourdb are provided in \cref{tab:ganit_difficulty_distribution}.

\section{Training \ours} \label{sec:ganitllm}
Following the success of GRPO \citep{shao2024deepseekmath, dipta2026pa3} in training LLMs for reasoning, tool-calling, and math, we use GRPO to train our model.
The total reward $R$ is computed as:
\begin{equation*}
R = R_{\text{format}} + R_{\text{correctness}} + R_{\text{bengali}} \in [0, 4]
\end{equation*}
where $R_{\text{format}} \in \{0, 1\}$ checks output format, $R_{\text{correctness}} \in \{0, 1, 2\}$ rewards correct answers (with bonus for Bengali answers), and $R_{\text{bengali}} \in \{0, 1\}$ rewards sufficient ($\geq$ 80\%) Bengali reasoning. The 80\% threshold is intentionally set below 100\% to ensure that sufficient token budget remains for mathematical notation and formulas, which are inherently language-agnostic. This threshold was empirically determined based on the distribution of Bengali tokens in the chain-of-thought traces available in our \oursft dataset. Furthermore, during reward computation, we strip punctuation from the token count to avoid artificially inflating or skewing the Bengali ratio score.
From initial runs, we had the following observations:

\begin{enumerate}[leftmargin=\parindent, itemsep=0.1em, topsep=0.2em]
    \item The policy model, i.e., \texttt{Qwen3}, tends to reason in English and then produce the Bengali answer, even when explicitly prompted to reason and answer in Bengali (\cref{app:prompt}). We hypothesize that this behavior stems from the predominance of English reasoning data during pre-training.

    \item Under standard shuffle-based GRPO training, the policy model fails to generate any correct answers within the rollout group, causing all advantage values to collapse to zero. As a result, the model fails to learn effectively.

    \item Many of the early generations are truncated due to the maximum token limit we imposed. These truncated outputs negatively impact the learning process. While increasing the token limit is possible, we observed that longer generations often contain repetition or unnecessary reasoning.
\end{enumerate}

Building on these observations, we introduce a multi-stage training recipe. \textbf{In the first stage}, we leverage the CoT-SFT split of \ourtrain to teach the model to reason in Bengali using fewer tokens. \textbf{In the second stage}, we apply a modified GRPO training procedure (described in \cref{sec:curriculum_grpo}) using the RL split of \ourtrain, enabling the model to generalize its reasoning ability and effectively solve Bengali math problems. 
To further mitigate the challenges of overlength generations, we follow \citet{yu2025dapo} and incorporate an overlength filter and token-level loss into the GRPO training.

\subsection{\cgrpo} \label{sec:curriculum_grpo}

As discussed earlier, using a traditional training strategy with random shuffling can result in ``Hard'' or ``Olympiad'' level problems appearing early in the training process -- before the model has developed the ability to solve them, even with a high rollout of 8. This is expected, as the hard or olympiad problems are those that even the large models (e.g. \texttt{Qwen3-32B}) take 32 turns to get an accurate solution. In such cases, the model receives zero reward across the whole group, leading to ineffective updates and stagnation in learning.

To address this, and inspired by recent advances in curriculum-based learning \citep{hammoud_train_2025, chen_self-evolving_2025, gao_prompt_2025}, we propose \textbf{\cgrpo}, a modified data sampling strategy that orders training data based on pseudo-difficulty.

A naive approach would be to sort the entire dataset from easy to hard based on difficulty. However, this can lead to early overfitting on simpler problems, making it harder for the model to adapt to more challenging samples later in training. To mitigate this, we adopt a soft curriculum strategy, using the number of correct generations as a fine-grained difficulty signal (ranging from 1 to 32) rather than relying on coarse difficulty categories. This provides more precise control over difficulty-aware sampling and allows the model to gradually strengthen its reasoning capabilities through increasingly difficult examples.

Specifically, for every bucket (1--32), we sample 60\% of examples (136 instances) from the current bucket and 40\% from the remaining 31 buckets (3 instances per bucket, totaling 93), resulting in 229 examples per bucket. We chose a 60/40 split based on preliminary experiments showing that higher primary-bucket ratios (e.g., 80/20) led to catastrophic forgetting of easier problems, while lower ratios (e.g., 50/50) diluted the curriculum signal. The 60/40 balance empirically provided stable training while maintaining sufficient difficulty progression.
Finally, we sort the training data by the primary bucket's difficulty level to ensure a smooth curriculum progression from easy to hard, reducing the risk of premature convergence or reward sparsity.
The full procedure is detailed in \cref{alg:curriculum-shuffle}.

\input{tables/alg_cgrpo}

%% file: tables/fig_data_processing.tex
\begin{figure*}[!t]
    \centering
    \includegraphics[width=0.95\linewidth]{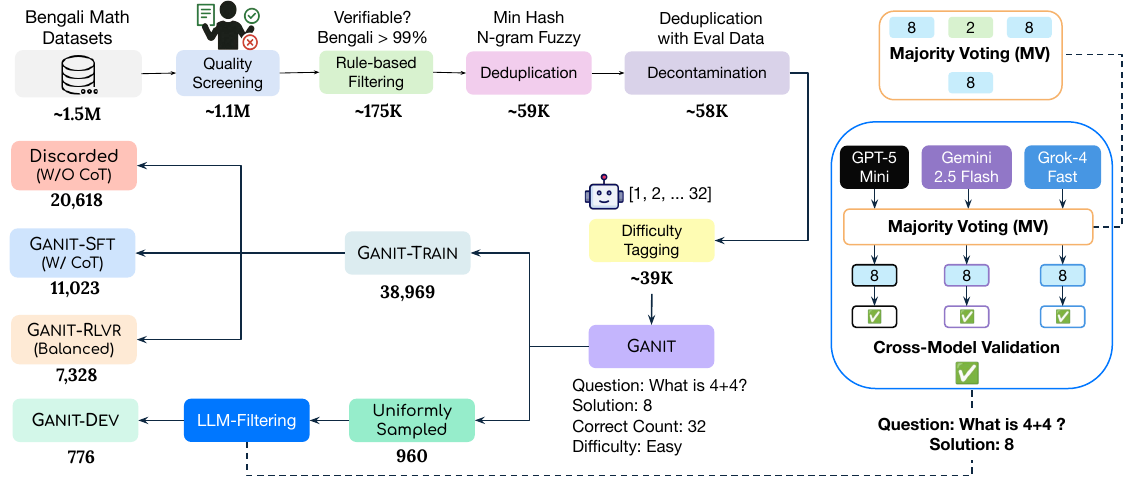}
    \caption{Overview of the \ourdb construction pipeline. Starting from $\sim$1.5M Bengali math problems, we apply multi-stage quality filtration, verification, deduplication, and decontamination to obtain \ourtrain (SFT and RLVR) and \ourdev.}
    \label{fig:data_preparation}
    \vspace{-4mm}
\end{figure*}

%% file: tables/alg_cgrpo.tex
\begin{algorithm}[t]
\caption{Curriculum-based dataset shuffling. Given a dataset $\mathcal{D}$, the algorithm outputs a curriculum-ordered version $\mathcal{D}'$, where each datapoint is tagged with a correctness count ranging from 1 (hardest) to 32 (easiest).}

\label{alg:curriculum-shuffle}
\begin{algorithmic}[1]
\Require Dataset $\mathcal{D} = \{(x_i, y_i, d_i)\}_{i=1}^N$ where $d_i \in [1, 32]$ is difficulty
\Ensure Curriculum-ordered dataset $\mathcal{D}'$
\State Group samples by difficulty: $\mathcal{D}_d \leftarrow \{(x_i, y_i) : d_i = d\}$ for $d = 1, \ldots, 32$
\State $n_d \leftarrow |\mathcal{D}_d|$ \Comment{Total Samples per difficulty}
\State $n_p \leftarrow \lfloor n_d \times 0.6 \rfloor$ \Comment{60\% Primary samples}%
\State $n_o \leftarrow \lfloor n_d \times 0.4 \rfloor$ \Comment{40\% Other Samples}%

\State $\mathcal{D}' \leftarrow \emptyset$
\For{$d_p = 32$ \textbf{down to} $1$} \Comment{Easy to Hard }
    \State $\mathcal{B} \leftarrow \mathcal{D}_{d_p}[1:n_p]$ \Comment{Primary samples}
    
    \For{each $d_o \neq d_p$}
        \State $\mathcal{B} \leftarrow \mathcal{B} \cup \mathcal{D}_{d_o}[\text{slice}]$ \Comment{Add $n_o$ samples}
    \EndFor
    
    \State Shuffle $\mathcal{B}$ randomly
    \State $\mathcal{D}' \leftarrow \mathcal{D}' \cup \mathcal{B}$
\EndFor

\State \Return $\mathcal{D}'$
\end{algorithmic}
\end{algorithm}

%% file: sections/03_results.tex
\section{Experiment Setup}

\paragraph{Datasets}
For SFT and GRPO training, we use \oursft and \ourrlvr, respectively. For evaluation, we use Bn-MGSM \citep{shi2022language} and Bn-MSVAMP \citep{chen2023breaking} benchmarks. Additionally, we report performance on our proposed \ourdev set.

\paragraph{Implementation Details}
We trained our model for 50 epochs during SFT and 5 epochs during GRPO. In both stages, the best checkpoint was selected based on accuracy on \ourdev. We used full fine-tuning for SFT and LoRA-based fine-tuning \citep{hu_lora_2021} for GRPO \citep{shao2024deepseekmath}. All training and inference were conducted on 2×A100 GPUs.
As base models, we used \texttt{Qwen3-0.6B}, \texttt{Qwen3-1.7B}, and \texttt{Qwen3-4B} \citep{yang_qwen3_2025}. Additional implementation details are provided in \cref{app:imple}. For evaluation, we used a temperature of 0.0 to ensure deterministic reproducibility, and applied the same prompt across all models (see \cref{app:prompt}).

\paragraph{Baselines}
While the primary goal of our paper is to develop a small Math LLM suitable for developing and low-resource countries, we also provide comparisons with larger models to compare our results. Specifically, we compare against the \texttt{Qwen3} family, ranging from 0.6B to 32B parameters.\footnote{We focus on a single model family to isolate the effect of our training recipe from base model differences. Additionally, no existing Bengali math reasoning models are publicly available for direct comparison.} For broader context, we also include results from \texttt{gpt-4.1}, \texttt{gpt-4.1-mini}, and \texttt{TigerLLM-9B}.

\section{Results \& Analysis}
The results on MGSM and MSVAMP are reported in \cref{tab:external_result}. In addition to accuracy, we report the number of generated words and the percentage of Bengali characters. Our goal is not only to develop a Bengali math LLM that produces correct answers, but also one that reasons in Bengali while maintaining a reasonable output length. This ensures both interpretability and practical usability for end users with limited computational resources.
Due to significant differences in tokenization between English and Bengali, we report word counts rather than token counts for fair comparison. To compute Bengali reasoning percentage, we count all characters in the model's reasoning output (excluding whitespace) and calculate what fraction belong to the Bengali Unicode block (\texttt{U+0980–U+09FF}). For instance, if a reasoning block contains 800 Bengali characters and 200 English/numeric characters, the Bengali percentage is 80\%.

The results indicate that while base models perform reasonably well, their success often comes at the cost of longer generations and reasoning primarily in English. This is expected, as much of the reasoning-related pretraining data is in English \citep{chowdhery_palm_2022}, often with long reasoning traces \citep{guo2025deepseek}.

Training the model with \oursft improves both accuracy and interpretability, yielding better performance, fewer generated words, and reasoning in Bengali. Further improvements are observed with \cgrpo, which pushes the results even further in the desired direction.

Notably, our 4B model outperforms the 8B base model and shows comparable results with the 14B model while being 2x and 3.5x smaller. Beyond accuracy, our approach yields dramatically more concise responses, \ours-4B generates answers averaging only 193 words compared to 943 for the base model, \textbf{a 79.5\% reduction}. More importantly, \textbf{Bengali characters increase from 14.79\% to 88.71\%}, indicating that our models reason in the target language rather than defaulting to English, addressing a critical limitation of Mathematical Reasoning.
While improvements remain consistent, the absolute gains from \cgrpo narrow at the 0.6B scale, suggesting that model capacity imposes a floor on achievable performance for complex mathematical reasoning. Also, previous studies \citep{nimmaturi_predictive_2025} have shown that a sufficiently capable model is needed to improve on reasoning with GRPO. We also provide a qualitative analysis in \cref{app:qual_analysis}.

\input{tables/main_results}

\section{Ablation Study}
\subsection{Impact of Multi-stage Training}
As the base-instruct models already show strong instruction-following capabilities, an intuitive question arises: \textit{Do we really need multi-stage training?} Or \textit{could we do the SFT/RL stage directly on top of the base-instruct model?}
Table \ref{tab:ab_multi_stage} presents an ablation study showing the necessity of our multi-stage training pipeline. We compare three training configurations: SFT only, \cgrpo (CGRPO) only, and our full pipeline (\ours{} = SFT followed by CGRPO).

\textbf{SFT establishes language grounding but provides limited reasoning gains.} SFT alone increases Bengali character usage (from 14.79\% to 86.65\% for the 4B model) while reducing the number of tokens. However, accuracy improvements are modest compared to CGRPO-based training.

\textbf{\cgrpo alone improves accuracy but sacrifices interpretability}. Applying CGRPO directly to the base model yields the highest raw accuracy, occasionally surpassing the 14B counterpart. However, this configuration retains only 14.94\% Bengali in reasoning, indicating that the model primarily reasons in English. This defeats the purpose of developing a Bengali mathematical reasoning system, as users receive little interpretable reasoning in their native language.

\textbf{Multi-stage training achieves the best trade-off.} Our sequential approach: first grounding the reasoning in Bengali through SFT, then enhancing reasoning via CGRPO—yields strong accuracy while maintaining high language adherence (88.71\% Bengali characters) and concise outputs (193 words). This demonstrates that the two stages serve complementary roles that cannot be achieved through single-stage training alone.

\subsection{Impact of \cgrpo} \label{sec:impact_cgrpo}
\input{tables/ab_multi_stage}
\cref{tab:ab_cgrpo} presents an ablation comparing standard GRPO against our proposed \cgrpo (CGRPO) within the multi-stage training pipeline. While both approaches achieve comparable final performance, CGRPO demonstrates substantially improved training efficiency.

\textbf{Comparable accuracy with dramatically faster convergence.} Across all model scales, GRPO and CGRPO achieve nearly identical accuracy on both benchmarks. For the 4B model, the difference is within 1 percentage point (77.60 vs. 76.80 on MGSM; 76.30 vs. 76.40 on MSVAMP). However, CGRPO reaches its optimal checkpoint at step 600 compared to step 2300 for vanilla GRPO -- a 3.8× reduction in training steps. This efficiency gap widens at smaller scales: for the 0.6B model, CGRPO converges at step 1300 versus 7300 for GRPO, representing a 5.6× speedup.

\textbf{Addressing the cold start problem.} The efficiency gains stem from CGRPO's curriculum-based sample ordering. In vanilla GRPO, random shuffling exposes the model to difficult examples early in training when it lacks sufficient capability, resulting in predominantly incorrect generations that provide weak or no learning signals at all (i.e. all rewards 0 across the group). This cold start problem delays meaningful policy improvement. On the other hand, CGRPO orders training samples by difficulty, allowing the model to first build foundational reasoning patterns before tackling complex examples.

\input{tables/ab_cgrpo}

%% file: tables/main_results.tex
\begin{table}[!t]
    \centering
    \definecolor{highlightcol}{RGB}{230, 242, 255}
    \resizebox{\columnwidth}{!}{%
    \begin{tabular}{@{}l|cc|cc@{}}
        \toprule
        & \textbf{\textit{MGSM $\uparrow$}} 
        & \textbf{\textit{MSVAMP $\uparrow$}} 
        & \textbf{\textit{\makecell{Words $\downarrow$}}} 
        & \textbf{\textit{\makecell{Bn (\%) $\uparrow$}}} \\
        \midrule

        gpt4.1 & \textbf{89.20} & \textbf{82.30} & \textbf{200} & 88.16 \\
        gpt4.1-mini & 87.20 & 78.60 & 232 & 88.18 \\
        TigerLLM-9B & 47.20 & 40.40 & 206 & \textbf{93.69} \\

        Qwen3-32B & {85.60} & {76.10} & {712} & {21.08} \\
        Qwen3-14B & 83.60 & 75.80 & 767 & 17.87 \\
        Qwen3-8B & 75.12 & 72.42 & 846 & 16.48 \\
        
        \midrule
        \midrule
        Qwen3-4B & 69.20 & 70.50 & 943 & 14.79 \\
        \ours-4B & \textbf{76.80} & \textbf{76.40} & \textbf{193} & \textbf{88.71} \\
        
        \midrule
        \midrule
        Qwen3-1.7B & 15.20 & 14.10 & 1124 & 19.64 \\
        \ours-1.7B & \textbf{52.80} & \textbf{66.80} & \textbf{210} & \textbf{87.80} \\
        
        \midrule
        \midrule
        Qwen3-0.6B & 8.40 & 12.20 & 1265 & 12.43 \\
        \ours-0.6B & \textbf{28.40} & \textbf{52.40} & \textbf{248} & \textbf{88.70} \\
        \bottomrule
    \end{tabular}%
    }
    \caption{Results on Bn-MGSM and Bn-MSVAMP test sets. \textbf{\ours enables smaller models to match or exceed larger counterparts}: \ours-4B surpasses Qwen3-8B by 7.6 points on MGSM while improving Bengali characters from 14.79\% to 88.71\%. \textbf{Bold} denotes best performance within each parameter category. Full results are provided in \cref{tab:full_main_result}.}
    \label{tab:external_result}
    \vspace{-4mm}
\end{table}

%% file: tables/ab_multi_stage.tex
\begin{table}[!t]
    \centering
    \definecolor{highlightcol}{RGB}{230, 242, 255}
    \resizebox{\columnwidth}{!}{%
    \begin{tabular}{@{}l|cc|cc@{}}
        \toprule
        & \textbf{\textit{MGSM $\uparrow$}} 
        & \textbf{\textit{MSVAMP $\uparrow$}} 
        & \textbf{\textit{\makecell{Words $\downarrow$}}} 
        & \textbf{\textit{\makecell{Bn (\%) $\uparrow$}}} \\
        \midrule
        Qwen3-4B & 69.20 & 70.50 & 943 & 14.79 \\
        \qquad + SFT & 74.00 & 74.60 & 184 & 86.65 \\
        \qquad + CGRPO & 82.40 & 78.50 & 844 & 14.94 \\
        \ours-4B & 76.80 & 76.40 & 193 & 88.71 \\
        
        \midrule
        \midrule
        Qwen3-1.7B & 15.20 & 14.10 & 1124 & 19.64 \\
        \qquad + SFT & 48.80 & 64.60 & 253 & 87.79 \\
        \qquad + CGRPO & 59.60 & 66.20 & 1002 & 18.74 \\
        \ours-1.7B & 52.80 & 66.80 & 210 & 87.80 \\
        
        \midrule
        \midrule
        Qwen3-0.6B & 8.40 & 12.20 & 1265 & 12.43 \\
        \qquad + SFT & 28.40 & 51.40 & 263 & 88.60 \\
        \qquad + CGRPO & 17.20 & 35.20 & 824 & 11.67 \\
        \ours-0.6B & 28.40 & 52.40 & 248 & 88.70 \\
        \bottomrule
    \end{tabular}%
    }
    \caption{Ablation study on multi-stage training. \cgrpo alone achieves competitive accuracy but fails to maintain Bengali language reasoning, while SFT alone provides limited reasoning improvements. \textbf{Our multi-stage pipeline combines the complementary benefits of both approaches.}}
    \label{tab:ab_multi_stage}
    \vspace{-4mm}
\end{table}

%% file: tables/ab_cgrpo.tex
\begin{table}[!t]
    \centering
    \definecolor{highlightcol}{RGB}{230, 242, 255}
    \resizebox{\columnwidth}{!}{%
    \begin{tabular}{@{}l|cc|cc|c@{}}
        \toprule
        & \textbf{\textit{MGSM $\uparrow$}} 
        & \textbf{\textit{MSVAMP $\uparrow$}} 
        & \textbf{\textit{\makecell{Words $\downarrow$}}} 
        & \textbf{\textit{\makecell{Bn (\%) $\uparrow$}}}
        & \textbf{\textit{\makecell{Best Ckpt. $\downarrow$}}} \\
        
        \midrule
        Qwen3-4B & 69.20 & 70.50 & 943 & 14.79 & - \\
        \; SFT + GRPO & 77.60 & 76.30 & 189 & 88.61 & 2300 \\
        \; SFT + CGRPO & 76.80 & 76.40 & 193 & 88.71 & 600 \\
        
        \midrule
        \midrule
        Qwen3-1.7B & 15.20 & 14.10 & 1124 & 19.64 & - \\
        \; SFT + GRPO & 53.60 & 66.90 & 207 & 88.32 & 7900 \\
        \; SFT + CGRPO & 52.80 & 66.80 & 210 & 87.80 & 2100 \\
        
        \midrule
        \midrule
        Qwen3-0.6B & 8.40 & 12.20 & 1265 & 12.43 \\
        \; SFT + GRPO & 32.40 & 52.50 & 246 & 88.45 & 7300 \\
        \; SFT + CGRPO & 28.40 & 52.40 & 248 & 88.70 & 1300 \\
        \bottomrule
    \end{tabular}%
    }
    \caption{Ablation study comparing GRPO and CGRPO. Both methods achieve comparable accuracy, but \textbf{CGRPO reaches optimal performance 3.8-5.6× faster} by addressing the cold start problem through curriculum-based sample ordering. Best Checkpoint denotes the training step at which peak validation performance was achieved.}
    \label{tab:ab_cgrpo}
    \vspace{-4mm}
\end{table}

%% file: sections/appendix/grpo_vs_cgrpo.tex
\begin{figure*}[!t]
\includegraphics[width=0.95\linewidth]{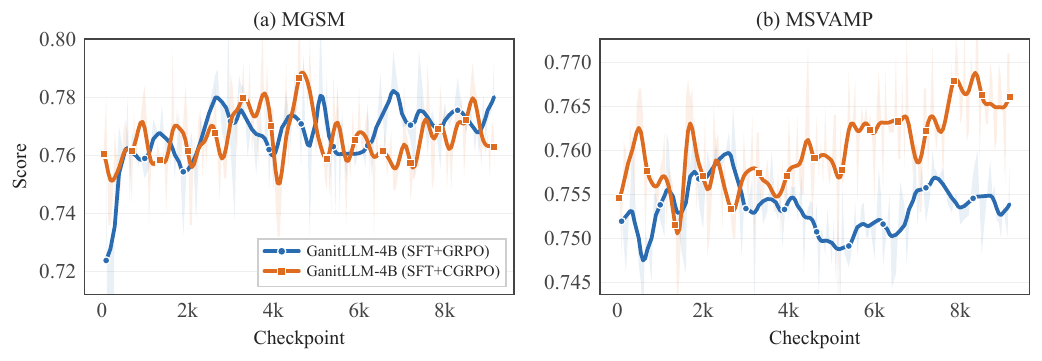}
\vspace{-4mm}
\caption{Evaluation curves comparing GRPO and \cgrpo on MGSM and MSVAMP benchmarks. Checkpoint-wise accuracy demonstrates that while both methods achieve comparable performance on the easier MGSM dataset (left), CGRPO substantially outperforms traditional GRPO on the harder MSVAMP dataset (right), where the cold-start problem causes GRPO to stagnate.}
\label{fig:grpo_vs_cgrpo}
\vspace{-4mm}
\end{figure*}

\section{Cold-Start Problem during GRPO} \label{app:grpo_vs_cgrpo}
We define the cold-start problem in GRPO as the scenario in which the policy model fails to generate any correct solution across the entire rollout group, resulting in zero gradients and suboptimal training. In \cref{fig:grpo_vs_cgrpo}, we plot checkpoint-wise accuracy on both MGSM and MSVAMP datasets under two configurations: (i) SFT $\rightarrow$ GRPO and (ii) SFT $\rightarrow$ \cgrpo.

The results show that on MGSM, the easier of the two datasets, traditional GRPO struggles initially but eventually catches up in accuracy after several hundred steps. In contrast, on MSVAMP, our proposed \cgrpo method demonstrates more efficient learning, exhibiting a steady upward trend in accuracy. Meanwhile, traditional GRPO stagnates, likely due to local optima caused by zero-gradient updates.

%% file: sections/appendix/qualitative_analysis.tex
\section{Qualitative Analysis} \label{app:qual_analysis}

In \cref{fig:qual_analysis}, we present model outputs for a representative Olympiad-level problem to illustrate the differences between training configurations. Results show that the base model (\texttt{Qwen3-4B}) produces correct answers but reasons primarily in English (7.58\% Bengali) with verbose outputs (932 words). SFT alone achieves high Bengali usage (97.63\%) and conciseness (645 words) but fails to produce the correct answer. \cgrpo alone improves accuracy but maintains English reasoning (7.32\% Bengali) and generates extremely verbose outputs (2223 words). Our full pipeline (SFT $\rightarrow$ \cgrpo) combines the benefits of both stages: achieving correct answers, native Bengali reasoning (97.7\%), and concise outputs (467 words), demonstrating that the two training stages serve complementary roles.

\subsection*{Key Findings}
\label{app:qualitative:findings}

\begin{enumerate}
    \item \textbf{SFT grounds language but not reasoning ability.} SFT alone successfully shifts the model's reasoning from English to Bengali (8\% $\rightarrow$ 97\% Bengali tokens) and reduces verbosity (932 $\rightarrow$ 645 words). However, it doesn't generate the accurate solution, suggesting that language grounding and mathematical reasoning are orthogonal capabilities.
    
    \item \textbf{GRPO improves accuracy but not language.} GRPO alone improves accuracy on difficult problems but the model continues to reason in English (8\% Bengali tokens).
    
    \item \textbf{Two-stage training combines both benefits.} Our SFT $\rightarrow$ GRPO pipeline produces outputs that are simultaneously accurate (matching GRPO-only), Bengali-dominant (matching SFT-only), and concise. This validates our hypothesis that SFT provides the language foundation that GRPO can then optimize without losing the language fidelity.
\end{enumerate}

%% file: sections/91_conclusion.tex
\section{Conclusion}
\label{sec:conclusion}

We address a critical gap in multilingual math reasoning: even when a base model can solve Bengali problems, it often reasons in English and merely translates the final answer. Our extensive ablations further show that traditional GRPO alone is inefficient for effective math reasoning in low-resource settings. To overcome these challenges, we first introduce \ourdb, a comprehensive, difficulty-aware Bengali math corpus with three splits: CoT-SFT, RLVR, and a validation set. Building on this resource, we then propose \cgrpo, a novel data-sampling strategy that significantly improves the efficiency of GRPO training for low-resource and underrepresented languages. Experiments show that our approach outperforms strong baselines in accuracy, token efficiency, and language fidelity, while converging faster than standard GRPO.

%% file: sections/99_limitations.tex
\section*{Limitations}
\label{sec:limitations}

While our method introduces a novel training paradigm to address the cold-start problem in low-resource settings, several limitations remain. 

First, our study is limited to a single language (Bengali) and a single domain (mathematical word problems). How well the proposed data construction and training recipe transfer to other low-resource languages remains an open research question.

Second, several components of our pipeline rely on proxy signals and automated tools. For instance, we tier difficulty using Pass@k scores from an evaluator model and filter development data using strong LLMs. Both can potentially introduce model-specific biases (e.g., systematically mischaracterizing certain problem types) that may propagate into training decisions.

Finally, our language-fidelity reward employs a character-percentage heuristic to approximate ``Bengali reasoning,'' which may incorrectly penalize valid outputs that mix languages, contain transliteration, or use symbols and numerals in ways that correlate with problem difficulty -- both of which the heuristic cannot distinguish from true code-switching.

\textit{Despite these limitations, we believe our work offers a strong first step toward grounded, language-consistent reasoning in low-resource settings and provides a practical, extensible training framework that can be adapted and refined in future research.}

%% file: sections/appendix/100_apendix.tex
\section*{Appendix} \label{sec:appendix}
\input{sections/appendix/code_switching}
\input{sections/appendix/full_result}

\input{sections/appendix/ganit_stat}

\input{sections/appendix/implementation}
\input{sections/appendix/mgsm_msvamp}

\input{sections/appendix/evaluator}

\input{sections/appendix/system_prompt}

\input{sections/appendix/ai_assistance}

\begin{figure*}
    \centering
    \includegraphics[width=1\linewidth]{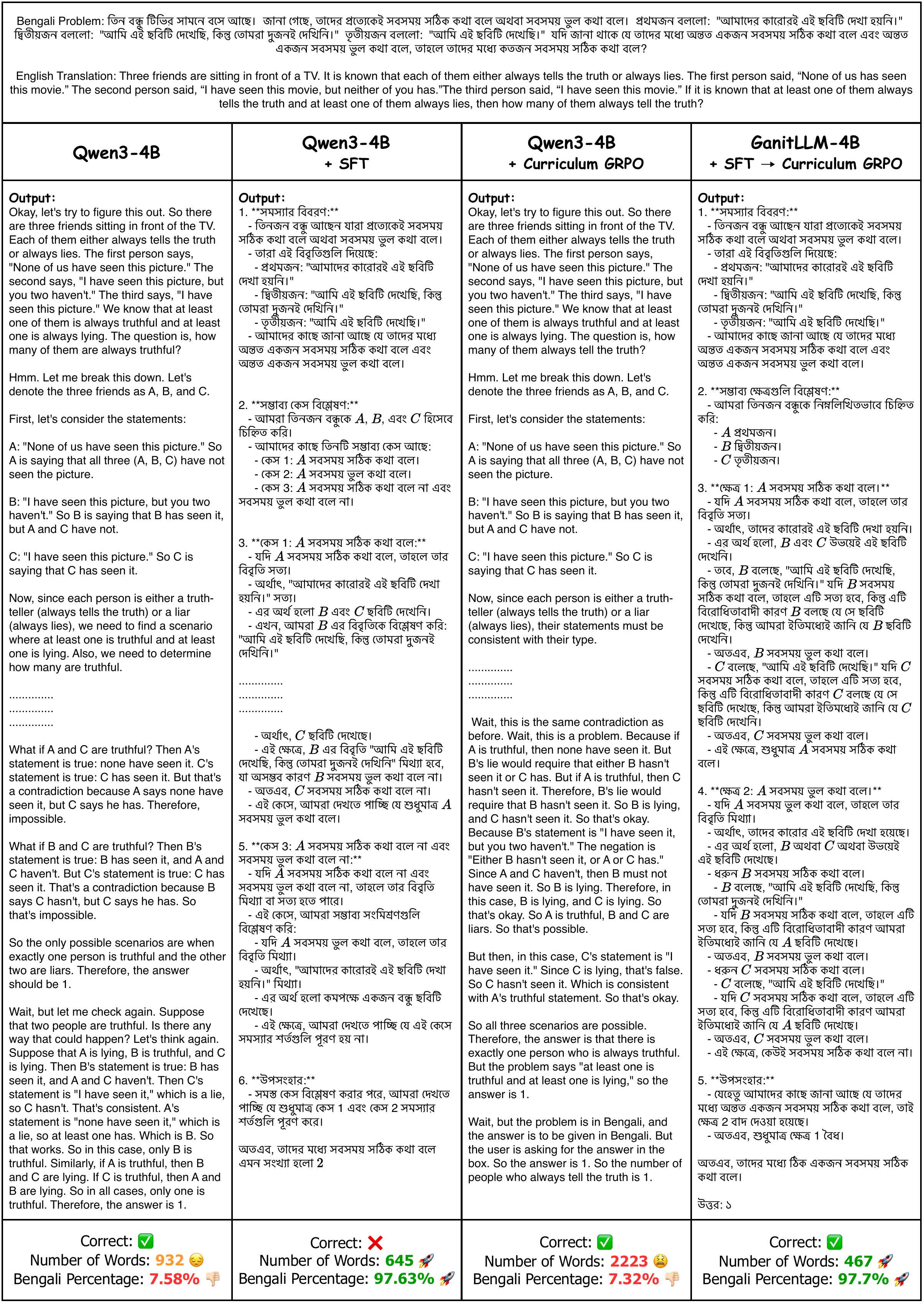}
    \caption{Qualitative comparison of training configurations on an \textbf{Olympiad-level} problem from \ourdev.}
    \label{fig:qual_analysis}
\end{figure*}

%% file: sections/appendix/code_switching.tex
\section{Code-Switching Analysis of \ours} \label{app:code-switching}

To quantify the extent of code-switching in our model's outputs, we conduct a fine-grained token-level analysis across all 1,236 \ours-4B generations on Bn-MGSM and Bn-MSVAMP. We classify each non-Bengali token into one of five categories and compute the code-switch rate as the number of Bengali$\leftrightarrow$English word-level transitions per sentence, excluding mathematical notation.

\begin{table*}[h]
\centering
\small
\begin{tabular}{lr}
\toprule
\textbf{Metric} & \textbf{Value} \\
\midrule
Bengali characters & 88.71\% \\
\midrule
\multicolumn{2}{l}{\textit{Non-Bengali token breakdown:}} \\
\quad Math (symbols, expressions, variables) & 47.55\% \\
\quad Numerals (incl.\ currency, percentages) & 34.17\% \\
\quad Markdown / formatting & 11.20\% \\
\quad Units / compounds (e.g., miles/hours) & 5.32\% \\
\quad \textbf{English words} & \textbf{0.69\%} \\
\midrule
Avg.\ code-switches per sentence & 0.014 \\
English word rate (of all tokens) & 0.11\% \\
Top English words & \texttt{of, gb, time, dvd, tv} \\
\bottomrule
\end{tabular}
\caption{Code-switching analysis of \ours-4B generations across Bn-MGSM and Bn-MSVAMP (1,236 samples; 101,869 tokens).}
\label{tab:code-switching}
\end{table*}

As shown in \cref{tab:code-switching}, only 115 out of 101,869 tokens (0.11\%) are actual English words, and these are predominantly domain-specific nouns carried over from the problem statements (e.g., ``dvd,'' ``tv,'' ``iphone'') rather than instances of reasoning in English. With an average of just 0.014 code-switches per sentence, virtually every sentence produced by the model is monolingual Bengali. The non-Bengali content is overwhelmingly composed of mathematical notation and numerals, which are standard conventions even in human-written Bengali mathematics and do not constitute code-switching in any meaningful linguistic sense.

%% file: sections/appendix/full_result.tex
\section{Additional Results}
\input{tables/main_results_long}
\input{tables/additional_hard_results}
Full results across MGSM, MSVAMP and \ourdev are reported in \cref{tab:full_main_result}.

We have also used two out-of-domain datasets: PolyMath \citep{wang2025polymath} and MT-MATH100 \citep{LinguisticGeneralizabilityTestTime_2025}. The results in \cref{tab:additional_results} show that  performance drops as difficulty increases (80.0\% → 17.76\% across PolyMath tiers). We attribute this to two factors: (1) our training data targets grade-level reasoning, lacking harder problem distributions, and (2) \ours was trained to output direct numerical answers, whereas problems in PolyMath and MT-MATH100 often require equation-level or symbolic outputs. Both limitations call for more challenging, diversified training data, which we identify as future work.

Nevertheless, three findings support our core contribution of efficient GRPO training for low-resource languages:

\begin{itemize}
    \item \ours-4B-SFT+CGRPO (80.0\%) surpasses Qwen3-8B (70.4\%) and approaches Qwen3-14B (82.4\%) on PolyMath-low, matching models 2–4× its size.
    \item Bengali adherence is consistently high (75–87\%) across all difficulty levels, while Qwen3-4B drops to 19–24\% Bengali on harder tiers, essentially reasoning in English.
    \item GRPO improves over SFT even on harder benchmarks (+11.87 on MT-MATH100), validating our core contribution of effective Curriculum-GRPO training for low-resource language reasoning.
\end{itemize}

%% file: tables/main_results_long.tex
\begin{table*}[!t]
    \centering
    \small
    \resizebox{\textwidth}{!}{
    \begin{tabular}{@{}l|ccccccc|cc@{}}
        \toprule
        
         & \multicolumn{4}{c}{\textbf{\ourdev$\uparrow$}} 
         & \multirow{2}{*}{\textbf{\textit{MGSM $\uparrow$}}} 
         & \multirow{2}{*}{\textbf{\textit{MSVAMP $\uparrow$}}} 
         & \multirow{2}{*}{\textbf{\textit{\makecell{Avg. \\ Accuracy $\uparrow$}}}}
         & \multirow{2}{*}{\textbf{\textit{\makecell{Avg. \\ Words $\downarrow$}}}} 
         & \multirow{2}{*}{\textbf{\textit{\makecell{Bn (\%) $\uparrow$}}}} \\
         
        \cmidrule(lr){2-5} 
         & Easy  & Medium & Hard & Olympic & & & \\
        \midrule

        gpt-4.1 & 92.83 & 90.10 & 84.41 & 78.18 & 89.20 & 82.30 & 86.17 & 200 & 88.16 \\
        gpt-4.1-mini & 89.69 & 81.19 & 83.87 & 72.73 & 87.20 & 78.60 & 82.21 & 232 & 88.18 \\
        TigerLLM-9B & 45.74 & 39.11 & 31.72 & 31.52 & 47.20 & 40.40 & 39.28 & 206 & 93.69 \\
        
        Qwen3-32B & 86.55 & 84.65 & 79.57 & 65.45 & 85.60 & 76.10 & 79.65 & 712 & 21.08 \\
        
        Qwen3-14B & 83.41 & 81.19 & 75.81 & 69.09 & 83.60 & 75.80 & 78.15 & 767 & 17.87 \\
        
        Qwen3-8B & 77.58 & 72.40 & 63.79 & 56.00 & 75.12 & 72.42 & 69.55 & 846 & 16.48 \\
        \midrule
        \midrule
        Qwen3-4B & 74.89 & 68.32 & 60.22 & 53.33 & 69.20 & 70.50 & 66.08 & 943 & 14.79 \\
        \quad + SFT & 63.23 & 50.00 & 45.70 & 41.82 & 74.00 & 74.60 & 58.23 & 184 & 86.65 \\
        \quad + CGRPO  & 86.10 & 77.23 & 72.04 & 70.91 & 82.40 & 78.50 & 77.86 & 844 & 14.94 \\
        \quad + SFT + GRPO & 67.26 & 50.00 & 50.00 & 49.70 & 77.60 & 76.30 & 61.81 & 189 & 88.61 \\
        \quad + SFT + CGRPO & 69.06 & 51.49 & 53.76 & 47.88 & 76.80 & 76.40 & 62.56 & 193 & 88.71 \\
        
        \midrule
        \midrule
        Qwen3-1.7B & 25.56 & 30.69 & 22.04 & 15.76 & 15.20 & 14.10 & 20.56 & 1124 & 19.64 \\
        \quad + SFT & 30.94 & 28.22 & 19.89 & 15.76 & 48.80 & 64.60 & 34.70 & 253 & 87.79 \\
        \quad + CGRPO  & 56.05 & 53.96 & 46.24 & 41.82 & 59.60 & 66.20 & 53.98 & 1002 & 18.74 \\
        \quad + SFT + GRPO  & 33.63 & 30.69 & 20.97 & 24.85 & 53.60 & 66.90 & 38.44 & 207 & 88.32 \\
        \quad + SFT + CGRPO  & 36.32 & 27.72 & 19.35 & 21.82 & 52.80 & 66.80 & 37.47 & 210 & 87.80 \\
        
        \midrule
        \midrule
        Qwen3-0.6B & 7.17 & 6.44 & 6.45 & 4.85 & 8.40 & 12.20 & 7.59 & 1265 & 12.43 \\
        \quad + SFT & 11.66 & 7.92 & 5.91 & 12.12 & 28.40 & 51.40 & 19.57 & 263 & 88.60 \\
        \quad + CGRPO  & 13.45 & 11.39 & 12.37 & 9.70 & 17.20 & 35.20 & 16.55 & 824 & 11.67 \\
        \quad + SFT + GRPO & 14.35 & 9.41 & 9.14 & 8.48 & 32.40 & 52.50 & 21.05 & 246 & 88.45 \\
        \quad + SFT + CGRPO & 13.90 & 8.91 & 10.22 & 12.73 & 28.40 & 52.40 & 21.09 & 248 & 88.70 \\
        \bottomrule
    \end{tabular}
    }
    \caption{Model Performance on \ourdev, Bn-MGSM \citep{shi2022language} and Bn-MSVAMP \citep{chen2023breaking} test set.}
    \label{tab:full_main_result}
\end{table*}

%% file: tables/additional_hard_results.tex
\begin{table*}[!h]
\centering
\small
\begin{tabular}{lcccccc}
\toprule
\multirow{2}{*}{\textbf{Model}} & \multicolumn{2}{c}{\textbf{PolyMath-low}} & \multicolumn{2}{c}{\textbf{PolyMath-medium}} & \multicolumn{2}{c}{\textbf{MT-MATH100}} \\
\cmidrule(lr){2-3} \cmidrule(lr){4-5} \cmidrule(lr){6-7}
 & Score & \%Bn & Score & \%Bn & Score & \%Bn \\
\midrule
Qwen3-4B & 68.80 & 62.46 & \textbf{52.94} & 23.95 & \textbf{84.75} & 38.28 \\
\ours-4B + SFT & 77.60 & 85.53 & 15.69 & 74.89 & 59.32 & 76.78 \\
\ours-4B + SFT + GRPO & 73.60 & \textbf{87.83} & 17.12 & 77.18 & 67.46 & 80.62 \\
\ours-4B + SFT + CGRPO & \textbf{80.00} & {86.59} & {17.76} & \textbf{77.64} & {71.19} & \textbf{79.52} \\
\bottomrule
\end{tabular}
\caption{Results on PolyMath and MT-MATH100 benchmarks.}
\label{tab:additional_results}
\end{table*}

%% file: sections/appendix/ganit_stat.tex
\section{\ourdb Statistics} \label{app:ganit_stat}
\cref{tab:ganit_difficulty_distribution} shows the difficulty and source distribution of our proposed dataset, \ourdb.

\input{tables/ganit_dataset_stat}

%% file: tables/ganit_dataset_stat.tex
\begin{table*}[!ht]
    \centering
    \small
    \begin{tabular}{@{}l cccc|ccccc@{}}
        \toprule
        \ourdb
        & \multicolumn{4}{c}{\textit{Difficulty Distribution}} 
        & \multicolumn{5}{c}{\textit{Source Distribution}} \\
        \cmidrule(r){1-1} \cmidrule(lr){2-5} \cmidrule(l){6-10}
        \textbf{Split} 
        & Easy & Med. & Hard & Oly. 
        & Numina & Somadhan & mCot & BDMO & s1K \\
        \midrule
        \oursft & 10015 & 84 & 208 & 716 & 7827 & 3039 & 157 & -- & -- \\
        \ourrlvr  & 1832 & 1832 & 1832 & 1832 & 6558 & 462 & 271 & 30 & 7 \\
        \ourdev & 223 & 202 & 186 & 165 & 704 & 40 & 27 & 4 & 1 \\
        \bottomrule
    \end{tabular}
    \caption{Difficulty and source distribution of the \ourdb dataset. The splits contain 11,023 (\oursft), 7,328 (\ourrlvr), and 776 (\ourdev) samples respectively. BDMO stands for Bangladesh Math Olympiad.}
    \label{tab:ganit_difficulty_distribution}
    \vspace{-4mm}
\end{table*}

%% file: sections/appendix/implementation.tex
\section{Implementation Details} \label{app:imple}
Full details of the hyperparameters used in the SFT and GRPO training is provided in the \cref{tab:sft_hyperparameters,tab:hyperparameters}, respectively.

\begin{table*}[!t]
\centering
\small
\begin{minipage}[b]{0.48\textwidth}
    \centering
    \begin{tabular}{@{}lr@{}}
    \toprule
    \textbf{Hyperparameter} & \textbf{Value} \\
    \midrule
    \multicolumn{2}{@{}l}{\textit{\textbf{Training Configuration}}} \\
    Training epochs & 50 \\
    Global batch size & 64 \\
    Learning rate & $1 \times 10^{-6}$ \\
    Learning rate scheduler & Cosine with Min LR \\
    Minimum learning rate & $1 \times 10^{-7}$ \\
    Warmup ratio & 0.05 \\
    Gradient clipping norm & 1.0 \\
    \midrule
    \multicolumn{2}{@{}l}{\textit{\textbf{Model Configuration}}} \\
    Training type & Full fine-tuning \\
    Precision & bfloat16 \\
    Max sequence length & 4096 \\
    \bottomrule
    \end{tabular}
    \caption{Supervised Fine-tuning (SFT) Training Hyperparameters}
    \label{tab:sft_hyperparameters}
\end{minipage}
\hfill
\begin{minipage}[b]{0.48\textwidth}
    \centering
    \begin{tabular}{@{}lr@{}}
    \toprule
    \textbf{Hyperparameter} & \textbf{Value} \\
    \midrule
    \multicolumn{2}{@{}l}{\textit{\textbf{Optimization}}} \\
    Learning Rate & $1 \times 10^{-4}$ \\
    Learning Rate Scheduler & Cosine with Min LR \\
    Minimum Learning Rate & $1 \times 10^{-5}$ \\
    Warmup Ratio & 0.05 \\
    Training Epochs & 5 \\
    Global Batch Size & 64 \\
    Gradient Clipping & 1.0 \\
    \midrule
    \multicolumn{2}{@{}l}{\textit{\textbf{LoRA Configuration}}} \\
    LoRA Rank & 16 \\
    LoRA Alpha & 32 \\
    \midrule
    \multicolumn{2}{@{}l}{\textit{\textbf{GRPO Parameters}}} \\
    Temperature & 1.0 \\
    Beta (KL Regularization) & 0.1 \\
    Number of Rollout & 8 \\
    Max Model Length & 4096 \\
    Max Completion Length & 2500 \\
    Loss & DAPO \\
    Dynamic Sample & True \\
    Max Resample Times & 3 \\
    Epsilon High & 0.28 \\
    Epsilon Low & 0.20 \\
    \bottomrule
    \end{tabular}
    \caption{Hyperparameters for GRPO LoRA Fine-tuning}
    \label{tab:hyperparameters}
\end{minipage}%
\end{table*}

%% file: sections/appendix/mgsm_msvamp.tex
\section{Evaluating Bn-MGSM \& Bn-MSVAMP} \label{app:eval_mgsm_msvamp}
\cref{tab:eval_mgsm_msvamp} presents the zero-shot evaluation results of eight open-source LLMs on the Bengali splits of both datasets. Rather than using single-point accuracy, we compute Pass@4, as it aligns with our difficulty tagging procedure based on Pass@k. Based on these results, we select \texttt{Qwen3-32B} as the evaluator model in our difficulty tagging pipeline.

\begin{table}[!ht]
    \centering
    \resizebox{\columnwidth}{!}{
    \begin{tabular}{@{}lcc@{}}
        \toprule
        Model ID & Bn-MGSM $\uparrow$ & Bn-MSVAMP $\uparrow$ \\
        \midrule
        Qwen2.5-14B-Instruct & 79.60 & 77.70 \\
        Qwen2.5-32B-Instruct & 88.80 & 80.90 \\
        Qwen2.5-72B-Instruct & 88.40 & 81.00 \\
        \midrule
        Qwen3-8B & 88.40 & 82.10 \\
        Qwen3-14B & 87.60 & 83.30 \\
        Qwen3-32B & \textbf{92.00} & \textbf{83.80} \\
        \midrule
        gpt-oss-20B & 88.40 & 82.10 \\
        \midrule
        Llama-3.3-70B & 60.40 & 77.40 \\
        \bottomrule
    \end{tabular}
    }
    \caption{Zero-shot evaluation (Pass@4) of MGSM and MSVAMP using 8 recent open-source LLMs.}
    \label{tab:eval_mgsm_msvamp}
\end{table}

%% file: sections/appendix/evaluator.tex
\section{Evaluator Details} \label{app:evaluator}
Two of the authors conducted quality screening of the seed data. Both hold graduate-level degrees, ensuring strong domain expertise and analytical skills. Their native language is Bengali, and they also possess advanced proficiency in English, enabling accurate evaluation of bilingual and technical content. Additionally, their advanced mathematical background qualifies them to assess materials requiring precise reasoning and quantitative understanding.

%% file: sections/appendix/system_prompt.tex
\section{User Prompt} \label{app:prompt}
We use the following user prompt templates for all of our experiments without altering the system prompt. Since our primary comparison models (e.g., the Qwen3 family) are all thinking models, we use a consistent prompt template across training and evaluation to ensure a fair and controlled comparison. Only GPT and TigerLLM use the instruct-style prompt template, and those results are presented as secondary comparisons in the paper. The difference between the two prompt variants is minimal---the thinking model prompt omits explicit \texttt{<think>} tags since these models natively produce structured reasoning traces, while the instruct model prompt includes \texttt{<think>} and \texttt{<answer>} tags to elicit a similar reasoning format. We believe this slight variation does not significantly affect our core claims.

\begin{tcolorbox}[
    coltitle=white,
    colframe=black,
    colback=black!5!white,
    breakable,
    enhanced,
    fontupper=\footnotesize,
    fontlower=\footnotesize,
    fonttitle=\small,
    title={User Prompt for Training \& Evaluation}
]
\textbf{Thinking Model (Primary: Qwen3 family, our models)}
\begin{Verbatim}[breaklines=true,breaksymbol=,fontfamily=tt]
A conversation takes place between the user and the assistant. The user asks a question, and the assistant solves the problem. Please reason step by step in Bengali, and put your final answer in the <answer> </answer> tags.

Question: {{problem}}
\end{Verbatim}

\textbf{Instruct Model (Secondary: GPT, TigerLLM)}
\begin{Verbatim}[breaklines=true,breaksymbol=,fontfamily=tt]
A conversation takes place between the user and the assistant. The user asks a question, and the assistant solves the problem. Please reason step by step, and put your reasoning process and final answer in the <think> </think> and <answer> </answer> tags respectively.

Question: {{problem}}
\end{Verbatim}
\end{tcolorbox}

%% file: sections/appendix/ai_assistance.tex
\section{Use of AI Assistance}
The authors have used Cursor\footnote{\url{https://cursor.com}} during development and ChatGPT\footnote{\url{https://chatgpt.com/}} for proofreading and polishing the final writing. Content provided to those tools were original to the authors.